# Roof fall hazard detection with convolutional neural networks using transfer learning

Ergin Isleyen[1], Sebnem Duzgun[1], R. McKell Carter[2]

[1] Colorado School of Mines, [2] University of Colorado Boulder

## Abstract

Roof falls due to geological conditions are major safety hazards in mining and tunneling industries, causing lost work times, injuries, and fatalities. Several large-opening limestone mines in the Eastern and Midwestern United States have roof fall problems caused by high horizontal stresses. The typical hazard management approach for this type of roof fall hazard relies heavily on visual inspections and expert knowledge. In this study, we propose an artificial intelligence (AI) based system for the detection roof fall hazards caused by high horizontal stresses. We use images depicting hazardous and non-hazardous roof conditions to develop a convolutional neural network for autonomous detection of hazardous roof conditions. To compensate for limited input data, we utilize a transfer learning approach. In transfer learning, an already-trained network is used as a starting point for classification in a similar domain. Results confirm that this approach works well for classifying roof conditions as hazardous or safe, achieving a statistical accuracy of 86%. However, accuracy alone is not enough to ensure a reliable hazard management system. System constraints and reliability are improved when the features being used by the network are understood. Therefore, we used a deep learning interpretation technique called integrated gradients to identify the important geologic features in each image for prediction. The analysis of integrated gradients shows that the system mimics expert judgment on roof fall hazard detection. The system developed in this paper demonstrates the potential of deep learning in geological hazard management to complement human experts, and likely to become an essential part of autonomous tunneling operations in those cases where hazard identification heavily depends on expert knowledge. Also, AI-based systems reduce expert exposure to hazardous conditions. They could also reduce inter-expert variability in hazard assessment and prevent knowledge loss in the case of personnel changes and expert unavailability.

## 1. Introduction

Roof falls are defined as the fall of material from above in underground mining and tunneling operations due to geological and operational conditions. Geological conditions involve the presence and orientation of discontinuities, stresses around the opening, geo-mechanical properties of roof materials, strata thickness, and moisture content. Operational conditions include excavation method, locality, height, and width of the openings. Seasonal effects of temperature and humidity also impact roof fall hazards (Pappas and Mark, 2012). Roof falls are responsible for a significant portion of accidents in underground mines, causing fatalities, injuries, damages to equipment, and lost work times.

The U.S. Mine Safety and Health Administration (MSHA) collects data on mining accidents and injuries, under Part 50 of the United States Code of Federal Regulations (CFR). Figure 1 shows the annual numbers of reportable accidents along with the percentage of roof fall-related accidents in underground mining operations in the United States between 2008 – 2018.

The total number of accidents and the number of roof fall accidents have been decreasing steadily due to improved safety management in underground mines. However, among 21 accident classes defined by MSHA, roof fall accidents still account for a significant portion of the total number of accidents as it is shown in Figure 1. Roof falls are the cause of many injuries and even fatalities in underground mines.

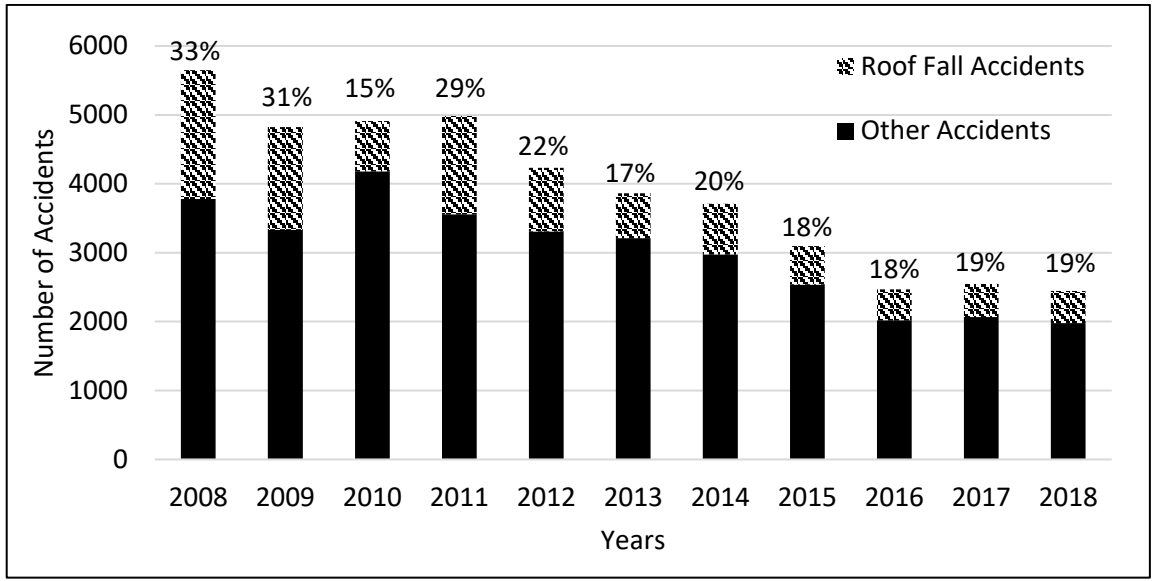

Figure 1. The total number of accidents along with the ratio of roof fall accidents to other accidents in the United States between 2008-2018 (MSHA Part 50 Data).

Various roof fall hazard management techniques have been implemented by the tunneling and the mining industries. Scaling and roof bolting are the two conventional roof fall hazard management techniques. In addition to these, many operations adopt more advanced technologies. Sherizadeh and Kulatilake (2016) used a numerical modeling technique to capture the geo-mechanical behavior of an underground mine with a roof instability problem and incorporated a roof fall detection criterion in their framework. Shen et al. (2018) established a numerical model to predict the displacement for the entry roof with a weak plane. Qin and Chian (2018) presented a kinematic analysis methodology for roof stability investigations in deep-buried tunnels. Shi et al. (2019) used a support vector machine method to predict deformation in shallow-buried tunnels. Sengani (2020) investigated the use of ground-penetrating radar to detect and distinguish seismic and non-seismic hazards in mining.

Recent advances in the field of artificial intelligence (AI) have created a promising alternative for developing fast and accurate hazard management tools. Tunneling and mining industries have started to use AI for various purposes in the last few years. Huang et al. (2018) proposed a novel image recognition algorithm based on deep learning to recognize crack and leakage defects of metro shield tunnels. Huang et al. (2018) used deep learning to identify the source location of micro-seismic events in underground mines. Xue et al. (2020) developed a deep learning-based method to automatically and accurately calculate the water leakage area in shield tunnel lining from images. Zhao et al. (2020) presented an image segmentation method for moisture marks of shield tunnel lining using convolutional neural networks. Wu et al. (2020) developed an AI system to monitor tunnel construction activities by integrating domain expertise to improve system accuracy. AI has not been used for hazard management in the tunneling and mining industries. However, developing an AI system requires a large number of data. In the case of roof fall hazard management in tunneling and mining, data collection needs to be done on-site. This may limit the number of data points since some areas may not be accessible for data collection due to safety concerns and operational restrictions.

Underground stone mines are prone to roof fall hazards as the openings are much taller compared to other underground mines. Iannacchione et al. (2005) investigated the use of microseismic monitoring to predict roof falls in stone mining. Although about 50% of the roof falls forecasted by this method were false alarms, predicted roof falls had an average warning time of 54 minutes. Iannacchione et al. (2006) proposed a Roof Fall Risk Index (RFRI) to quantify the roof fall hazards in underground stone mines using an observational technique. The RFRI consists of ten "defect" categories that include geology, mining-induced factors, roof profile, and moisture conditions. These categories were determined based on the extensive experience of the authors with underground stone mines, combined with an examination of the literature. Iannacchione et al. (2007) combined RFRI with a risk assessment technique. In this methodology, RFRI is used to identify potential roof fall hazards, which is used as a proxy for the probability of a roof fall occurring. The other parameters used in the calculation of the risk are the potential of a miner being injured and the severity of the roof fall event. Mine engineers can then display roof fall hazards on a risk map generated by this methodology. Bertoncini and Hinders (2010) used fuzzy classification to predict roof fall events in limestone mines using microseismic data. Bertoncini and Hinders' classifier successfully forecast two major roof fall events more than 15 hours before the first visual signs. Finally, in most underground stone mines, ground control personnel develop an intuitive judgment for the detection of roof fall hazards in their specific geological and operational domain. On a daily basis, these personnel use their intuitive judgments to make decisions regarding roof fall hazards.

For this reason, autonomous operations in underground spaces require robust geological hazard assessment methodologies. In geological hazard management, the development of AI-based hazard detection systems is of increasing importance. AI systems have the potential to support autonomy in hazard management for mining and tunneling operations and hence, allows elimination of expert personnel from exposing themselves to hazardous conditions during inspections. Also, in cases where hazard detection depends on intuitive judgments and visual inspections, knowledge transfer between personnel becomes a major challenge. Since the intuitive skills are learned by experience, new personnel training for hazard management takes extensive efforts and time, and when an experienced employee leaves, the knowledge may be lost. Exacerbating the loss, personnel and equipment become more vulnerable to safety risks during the training period of new personnel. An AI-based hazard detection system would mitigate this problem and reduce safety risks emerging during the absence of experienced ground control personnel. Finally, as the mining and tunneling industry transitions toward autonomous operations, hazard management tools that work continuously without a human in the loop become a necessity. In this study, we present an autonomous hazard management system using AI and demonstrate its implementation in Subtropolis Mine near Petersburg, Ohio. This mine experiences frequent roof falls and ground control personnel is responsible for doing daily inspections to detect roof fall hazards for a safe operation. The proposed autonomous hazard management system trains the last layer of a convolutional neural network (CNN) to capture the expert's intuitive judgment. In other words, the system aims to mimic the decision-making skills of a roof control expert for future roof-fall hazard detection without the expert`s presence. Most AI systems operate as black-box algorithms, meaning that the algorithm does not explain the logic behind its predictions. The lack of transparency of such models decreases the confidence of the users to utilize the models on critical tasks such as hazard management since the results of a roof fall can be catastrophic. To overcome this issue, in this study, we analyze the predictions of our model using an interpretation technique to build confidence in the AI-based roof fall hazard detection system.

## 2. Brief Description of the Roof Fall Problem in the Case Study

Horizontal stresses cause severe ground control problems in underground limestone and coal mines throughout the Eastern and Midwestern United States (Iannacchione et al., 2002; Mark and Mucho, 1994). The concentration of horizontal stresses in bedded deposits originates from plate tectonics. Stress measurements around the Eastern and Midwestern United States show that the direction of maximum horizontal compressive stress ranges from N 70°E to N 80°E in most mines. This direction is the same as the direction of movement as the North American Plate moves away from the Mid-Atlantic Ridge at a rate of ~2.5 cm/year (Figure 3) (Iannacchione et al., 2002).

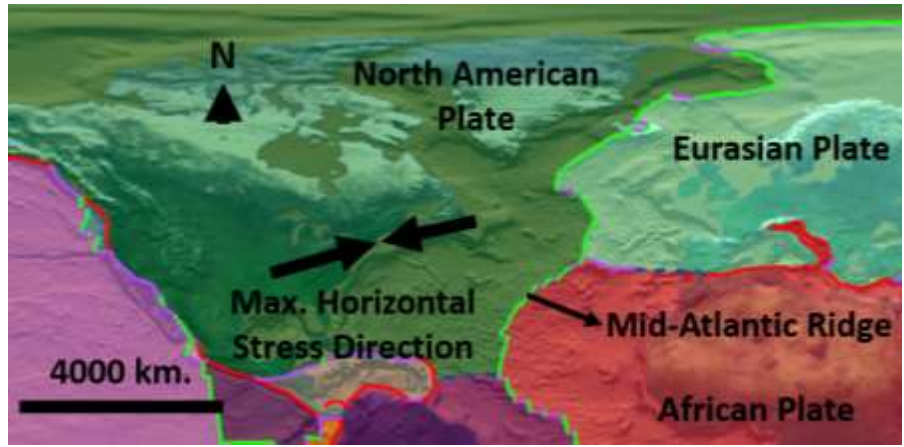

Figure 2. Effect of plate tectonics on horizontal stress on bedded deposits around Eastern and Midwestern United States (Adapted from Sokolowsky, 2004)

The tectonic stresses around the region build a constant strain field (between 0.00045 – 0.00090 $\varepsilon$), which induces higher levels of horizontal stress in soft limestone formations. This explains why some underground stone mines experience high levels of horizontal stress (Esterhuizen et al., 2008). Dolinar (2002) also shows that variations in the magnitude of the horizontal stresses in the Eastern and Midwestern United States are better explained by the elastic modulus of the rock, than by overburden depth.

The Subtropolis mine in eastern Ohio has been operating since 2006. The mining method is room and pillar, with rooms 9 – 12 m wide and 5 m high. The rectangular pillars have a length-to-width ratio ranging from 1.5 to 2, with lengths of 14 m and widths of 8 m. The mine has a history of ground control issues related to high horizontal stresses.

The mine layout has been developed to control high horizontal stress concentrations, with varying degrees of success. The advantage of adopting a stress control layout is that it maximizes the number of headings driven parallel to the direction of maximum stress, thereby reducing stress-related ground control problems. The Subtropolis Mine implemented a new stress control layout after experiencing frequent roof fall problems. Iannacchione et al. (2020) provided a thorough explanation of the mine layout considerations in the Subtropolis Mine. The headings advanced in East-West direction until 2007 when the orientation of the stress field was thought to be closer to North-South, and the Subtropolis mine layout was reoriented. A North-South mining orientation was then used until 2018 when the ground conditions worsened, and mining operations had to be paused in several faces. Horizontal stresses induce large oval-shaped roof falls, and the long-axis orientation of the ovals is generally at right angles to the direction of maximum horizontal compressive stress. In the Subtropolis mine, this orientation was found to be around N 55°E. Therefore, the direction of horizontal stress should be N 35°W, which was confirmed by an investigation of strata within roof exploration boreholes. Following this determination, all new headings have been aligned N 35°W. With this new mine layout, ground control issues have been transferred to the crosscuts, which mine management expected.

Subtropolis ground control experts have been able to associate areas of high horizontal stress with roof beams, both in entries and crosscuts. The roof beams are used as indicators of hazardous roof conditions (Figure 4).

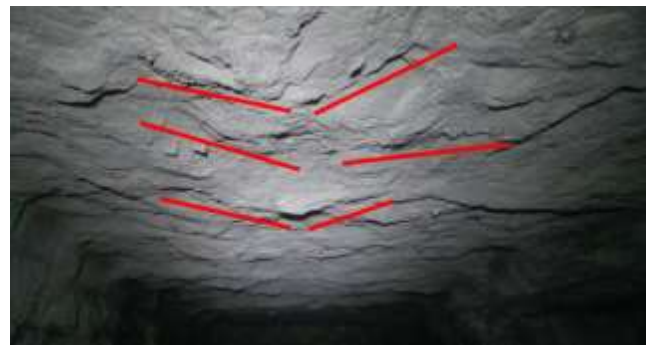

Figure 3. Roof beams on the mine roof

Figure 4 shows a hazardous roof condition in the Subtropolis Mine. Ground control personnel utilize the frequency and depth of the stress-induced roof beams in defining hazard levels. To keep track of the roof fall hazards in the mine and improve hazard management, mine personnel regularly map the roof beams. Figure 5 shows a sample hazard map.

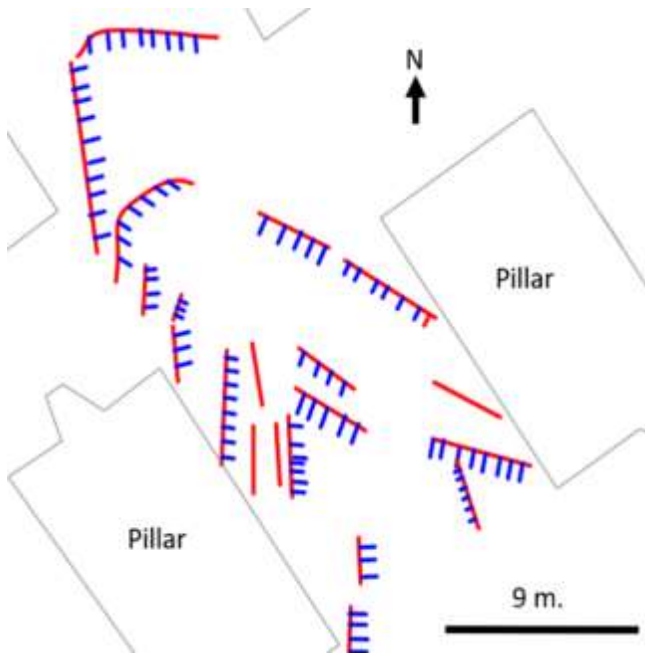

Figure 4. Roof beams are shown on the mine map

In Figure 5, red lines represent stress-induced roof beams. The blue lines drawn perpendicular to red lines represent the depth of a roof beam. The map is created by experienced ground control personnel based on their visual interpretation. The frequency and depth of these features can be interpreted using the mine map which enables the personnel to keep track of the hazardous areas.

### 3. METHODOLOGY

Since expert judgment depends on visual observations, we sought a system to work with similar visual inputs. We, therefore, chose a convolutional neural network (CNN) trained to recognize a variety of visual features that the rest of the model can use for classification or object detection. Figure 6 illustrates how CNN is integrated into the hazard identification and interpretation system.

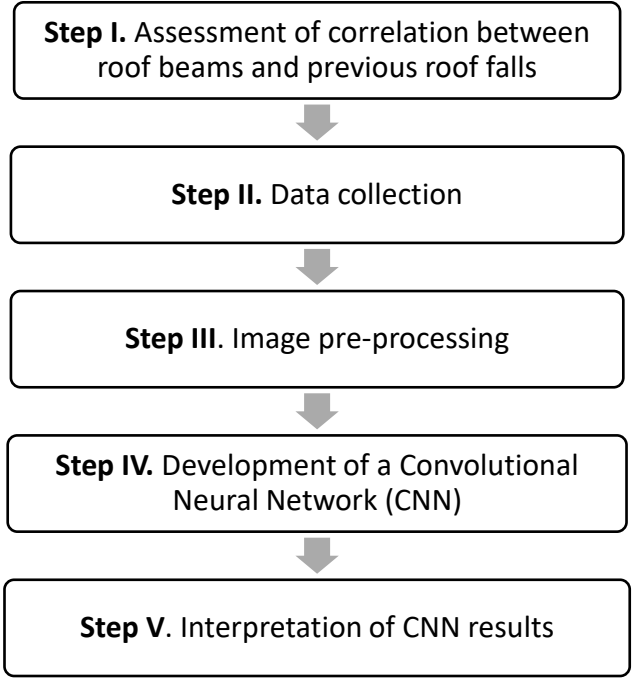

Figure 5. Research Methodology

In the first step, it is shown that the depth and frequency of roof beams correlate with the locations where roof falls occurred previously in the mine. This verifies the accuracy of the expert's description of how they identified the hazards, with quantitative methods. In the second step, the collection of image data that are used in developing the AI-based roof-fall-hazard detection system is explained. In the third step, images are processed to convert them into a suitable input format for the CNN algorithm. In the fourth step, the last layer of the CNN is retrained to classify images as hazardous or non-hazardous. The prior layers of the CNN were pre-trained with a large visual database. In the final step, a model interpretability algorithm is applied to the CNN model to recover features utilized by the CNN. The following explains the details of each step.

### 3.1. Assessment of Correlation between roof beams and previous roof falls

Ground control personnel at the Subtropolis Mine use their experience-based skills from years of mining experience in the region to identify roof fall hazards from stress-induced roof beams. They have identified the depth and the frequency as the most important characteristics of the stress-induced roof beams.

Therefore, they map the roof beams in a way that represents depth and frequency, see Figure 5. Statistical analysis demonstrates the correlation between the characteristics of the roof beams and previous roof falls.

Figure 7 shows the locations of 58 known roof falls provided by mine personnel. These were compared to 58 randomly selected locations with no previous roof falls.

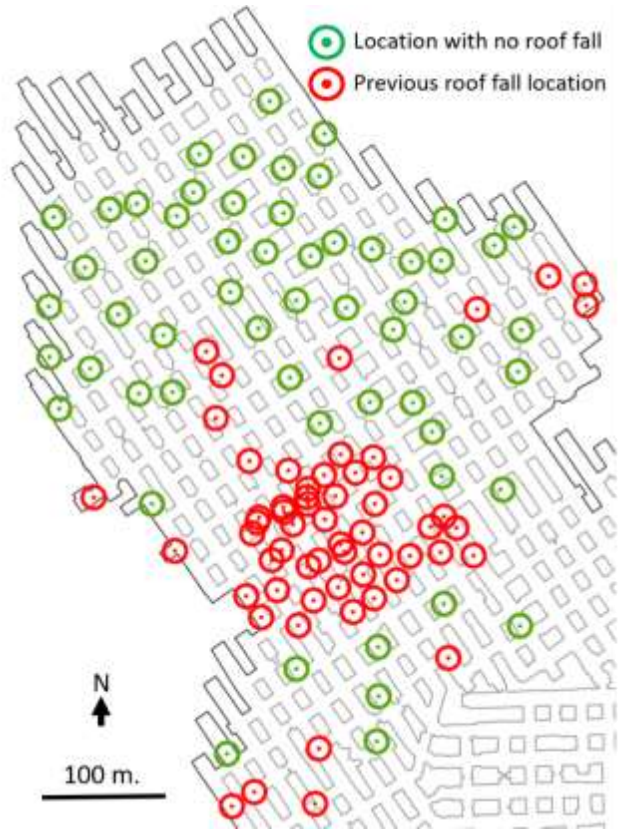

Figure 6. Previous roof fall and no-roof fall locations

To test for an association between roof beams and roof-fall hazards, a circle with a radius of 9 meters was drawn at each of the 116 locations and the number of roof beams inside each circle was recorded. The area covered by the circles match the size of the rooms, and it is assumed that this area is the effective area of a roof beam regarding roof falls. We also recorded the average depth of the roof beams within each circle. At previous roof fall locations, we used the roof beams mapped before the failure. Figure 8 shows an example of the depth and frequency of roof beams inside a circle centered on a previous roof-fall location.

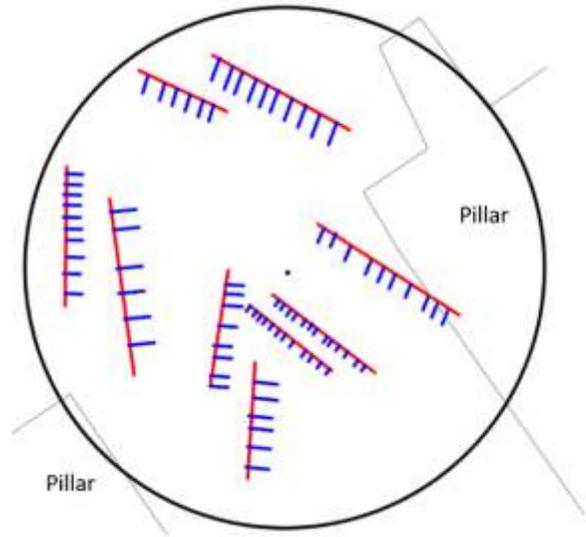

Figure 7. Roof beams around a previous roof fall location

In Figure 8, the frequency of roof beams is 9 (number of red lines which represent roof beams), and the average depth of roof beams is 2.2 in map units (the average length of blue lines which represent the depth of roof beams). Map units represent the length of the lines on the map and not the actual length measurements. Using this technique, frequency and depth information for each previous roof fall location and no-roof fall location is recorded. A statistical t-test was used to test for a difference between the means of the two groups for each feature. Summary statistics and the p-values are given in Table 1.

Table 1. Summary statistics (Mean, Standard Deviation (SD), Degrees of Freedom (df), t-values (t) and P-Values (P-val)

|  | Frequency | | Depth | |
|---|---|---|---|---|
|  | Roof fall | No - roof fall | Roof fall | No - roof fall |
| Mean | 9.76 | 7.05 | 1.47 | 1.17 |
| SD | 3.76 | 3.21 | 0.65 | 0.73 |
| df | **57** | | **57** | |
| t | **4.07** | | **2.28** | |
| P-val | **<.001** | | **.02** | |

For both frequency and depth values of roof beams, the P-values between "previous roof fall" and "no-roof fall" locations are smaller than the level of significance (0.05). We, therefore, conclude that the difference in frequency and depth between roof beams around previous roof fall locations and no-roof fall locations is unlikely to exist by chance. It is therefore plausible that the frequency and depth of roof beams can be used as indicators of a roof fall hazard.

This analysis verifies the success of ground control personnel's description of their intuitive decision-making skills in identifying features associated with roof falls. It, therefore, seems promising to target the identification of these features as an image recognition problem. In the next step, we describe image collection procedures at locations categorized as hazardous and non-hazardous by ground-control personnel.

### 3.2. Data Collection

Images were collected under two roof condition classes; hazardous and non-hazardous. These images were used to train and validate the convolutional neural network algorithm. The locations where the images are collected were determined based on the recommendations of the ground control experts at the mine site (Figure 9).

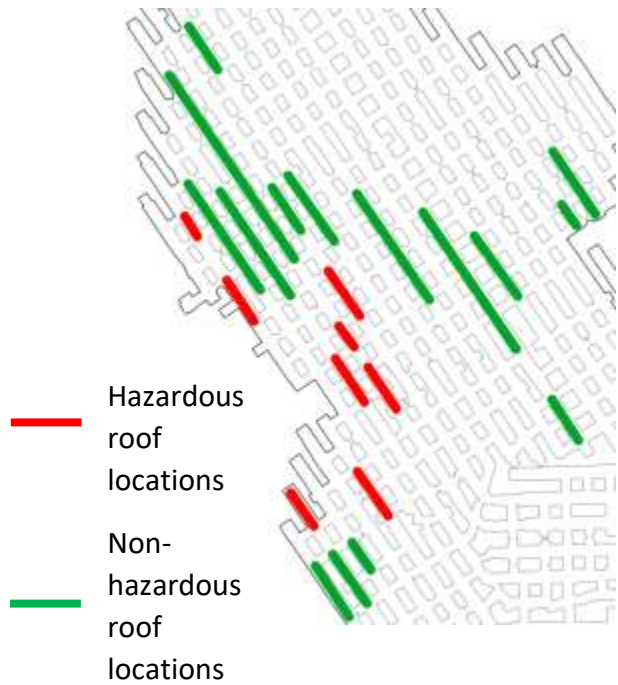

Figure 8. Data collection locations

The hazard level difference between locations suggested by the mine personnel is demonstrated by comparing the means of frequency and depth of roof beams between the two groups with a statistical t-test. The frequency and the average depth of roof beams are measured at 18 m long intervals that are marked for data collection. The results of the statistical analysis for the sites chosen are given in Table 2.

Table 2. Summary statistics (Mean, Standard Deviation (SD), Degrees of Freedom (df), t-values (t) and P-Values (P-val) for data collection locations

|  | Frequency | | Depth | |
| --- | --- | --- | --- | --- |
|  | Haz. | Non-haz. | Haz. | Non-haz. |
| Mean | 14.5 | 8.62 | 1.89 | 1.36 |
| SD | 20.4 | 11.1 | 0.43 | 0.28 |
| df | 51 | | 44 | |
| t | 4.46 | | 3.7 | |
| P-val | <.001 | | .01 | |

The p-values for frequency and depth are smaller than the level of significance (0.05). We, therefore, conclude that the difference between roof beams (in terms of frequency and depth) around expert-categorized hazardous and non-hazardous data collection locations is unlikely to have occurred by chance. It also makes it more likely that an image classifier could be used to index roof-fall risk.

Images were obtained using a Nikon D5000 camera, from locations labeled by human experts as hazardous and non-hazardous. To provide steady lighting during data collection, an LED light source was used. For hazardous and non-hazardous roof conditions, 83 and 166 images were collected, respectively. Figure 10 shows examples of hazardous and non-hazardous roof condition images.

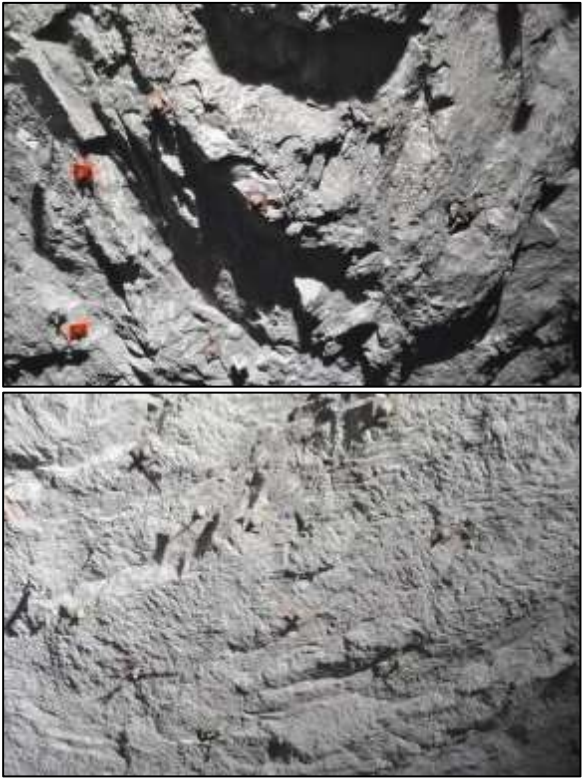

Figure 9. Example images; Hazardous roof (left), Non-hazardous roof (right).

### 3.3. Image Pre-Processing

To increase the number of available training and validation images, and to bring the images down to standard input size for the CNN, we first split the images into four tiles (Figure 11).

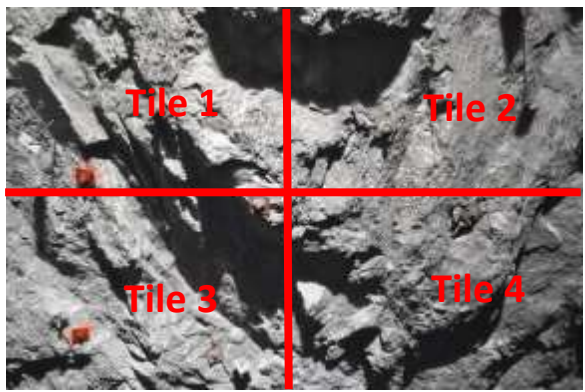

Figure 10. Image tiles

The images obtained after tiling goes through a data augmentation stage before used in CNN. The data augmentation method applied in this study rotates each image between 0° - 15°, randomly flips images horizontally, and randomly changes brightness, contrast, and saturation of the images. The data augmentation was only applied to training images, and not on validation images. In this way, the network processes different versions of the same images during the training and validates its accuracy on unaltered versions of the images. Different random transformations were applied at each epoch during the training. After data augmentation, we reduced the image size to 224 x 224 as required by the CNN.

### 3.4. Transferring a Convolutional Neural Network

Deep learning is the collective name for computational methods that transform raw input into a representation at a more abstract level with multiple layers of representation. For classification tasks, the input weights from layer to layer are learned to maximize the number of correctly classified images. Convolutional neural networks (CNN) are a particular deep learning architecture designed to learn to recognize recurring features. They capably process data that is in the form of multiple arrays, e.g. an image with three color channels. A CNN network typically represents small and meaningful, low-level features of an image, e.g. edges and dots, in the early layers. The later layers recognize objects as combinations of the low-level features (LeCun et al, 2015). Detecting low-level features for repeated use reduces the computational memory requirement and improves statistical efficiency (Bengio et al., 2017). CNN has been applied to object detection, segmentation, and recognition in images with great success. Some examples of CNN applications include face recognition (Taigman et al., 2014), pedestrian detection (Sermanet et al., 2014), and traffic sign recognition (Ciresan et al., 2012). Recent advances in CNN algorithms raised the possibility of using them to develop hazard detection systems. Perol et al. (2018) present an application of CNN for earthquake detection and location tracking from seismograms. Using CNN, Perol et al. were able to detect 17 times more earthquakes than had been cataloged by the Oklahoma Geological Survey. Yaloveha et al. (2019) used CNN to solve the problem of identifying forest areas with high fire hazards. The system uses multispectral images and spectral indices that detect vegetation and calculate moisture and carbon content. Wilkins et al. (2020) used CNN to detect microseismic events in coal mining operations and the results showed that the network's accuracy exceeds that of a human expert. Fang et al. (2020) developed hybrid models to improve the accuracy of landslide susceptibility predictions by integrating a CNN in their model.

Some CNN architectures focus on increasing network depth to improve classification accuracy. However, research has shown that deeper networks are harder to train, and network performance starts to degrade because of the vanishing gradient problem (Glorot and Bengio, 2010). He et al. (2015) addressed this degradation problem by introducing ResNet architecture. In ResNet, layers are reformulated as learning residual functions regarding the layer inputs, instead of learning unreferenced functions. He et al., (2015) presented evidence that these networks are easier to optimize and can gain accuracy with increased depth. In this study, a ResNet CNN architecture is used to develop the AI-based roof fall hazard detection system.

Fully training a CNN requires a large number of images. In this study, the number of images collected was small even after tiling the images to quadruple the number, and insufficient to train a network from scratch. We, therefore, take a different approach. Transfer learning has emerged as an alternative when insufficient data prevents achieving acceptable performance on classification tasks. In transfer learning, a network trained in one domain of interest is used for classification in another domain of interest with fewer training data (Pan and Yang, 2010). The idea behind transfer learning is that many networks learn low-level features that are not specific to any one image class and can, therefore, be used in other tasks. Practically, this means we can take an existing model to recognize novel image classes by retraining only the last layers where features are aggregated into objects. The transferability of features decreases as the similarity between the original task and target task decreases. However, Yosinski et al. (2014) showed that transferring features even from distant tasks can give better results than using random features. An example of using transfer learning in geosciences is presented by Li et al. (2017). They applied the technique to the classification of sandstone images. Cunha et al. (2020) used transfer learning to train an existing classifier to detect faults based on seismic data.

This study used a 152 layered ResNet network pre-trained on the ImageNet dataset. ImageNet is an image database that contains 1.2 million images with 1000 categories (Deng et al., 2009). The final fully connected layer of the network is reset and trained with the images collected for this study. The total number of parameters trained is then 525000. A validation dataset was created by separating the 20% of the collected images for hazardous and non-hazardous classes. Distribution of the total number of images used for training the fully connected layer and validation between both classes is given in Table 3.

Table 3. Distribution of number training and validation data between classes

|  | Hazardous | Non-hazardous | Total |
|---|---|---|---|
| **Training** | 266 | 532 | 798 |
| **Validation** | 66 | 132 | 198 |
| **Total** | 332 | 664 |  |

### 3.5. Interpretation of CNN Results

Deep learning models are typically "black boxes", meaning that the user does not receive a logical explanation for the predictions made by the model, regardless of how accurate the predictions are. However, if there are actions to be taken based on the predictions of the network, as in the case of roof fall hazard detection, understanding the reasons behind the predictions carries great importance for the user. Understanding the logic behind networks increases the confidence of the user, which is critical for managing hazards that could cause severe injuries and fatalities.

Since CNN takes images as input, the interpretability of CNN models can be achieved by producing visual explanations. Several techniques have been developed to interpret the predictions made by CNN. These techniques usually create visualizations that highlight image regions that contribute most to network predictions at the pixel level (Sundararajan et al., 2017; Fong et al., 2017; Selvaraju et al., 2017; Shrikumar et al., 2017) or object-level (Zhang et al., 2018). In this study, an attribution technique called "integrated gradients" is used for attributing the predictions of

CNN to the input images (Sundararajan et al., 2017). In this technique, a series of images are interpolated, increasing in intensity, between a black baseline image and the original image. The network predicts the class of each interpolated image and calculates the confidence scores of each prediction. The region is marked where the confidence scores keep increasing up to their final value. This region is where the important features of the original image are revealed by the increasing intensity of the interpolated images. Gradients of the interpolated images within that region are calculated, and the integration of these gradients gives the integrated gradients of the original image. The integrated gradients technique is pixel-based; therefore, the outputs show the most important pixels in each image for the hazard or non-hazard prediction.

## 4. RESULTS AND DISCUSSION

Roof-fall-hazard management at the Subtropolis Mine depends on the ground-control experts' visual interpretation of the roof beams formed by high horizontal stresses. Quantitative methods show a strong relationship between roof beams and roof fall incidents. This supports the experts' description of their intuitive decision-making skills in identifying features that show signs of hazardous roof conditions. It is therefore plausible to use expert-categorized hazardous and non-hazardous roof locations as a basis for image collection for the CNN image classifier which targets the detection of expert identified features.

The results of the CNN image classifier are interpreted using the validation accuracy and the training accuracy (Figure 12). The batch size used in this study is 16. The batch size is the number of images that propagates through the network at each iteration. Smaller batch sizes allow learning to start before the algorithm sees the majority of the data, which leads to faster convergence, whereas larger batch sizes cause significant loss of generalization ability of the model (Keskar et al., 2016). An epoch is when the entire dataset is passed forward and backward through the network once. Iterations are the number of batches needed to complete one epoch.

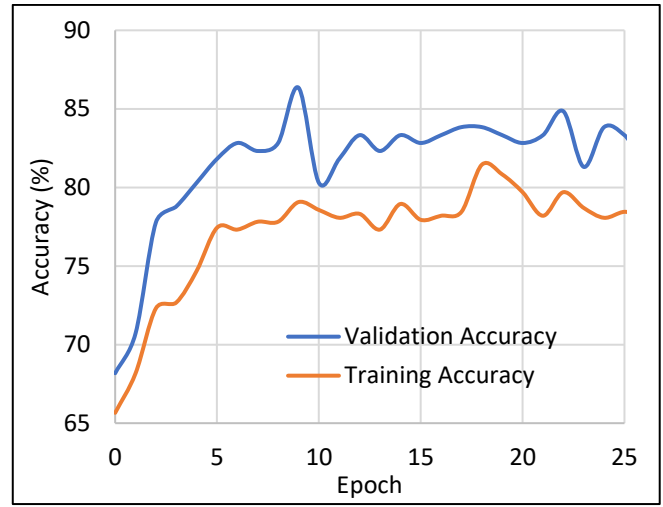

Figure 11. Validation and training accuracy of the network

The highest validation accuracy is 86.4%, reached at the 9$^{th}$ epoch. The out-of-sample validation dataset is used to obtain the validation accuracy of the network. In addition to validation accuracy, the training accuracy is calculated: this is the accuracy of the model when applied to the training images. Training accuracy is used as an indicator of overfitting which means that the model fits too closely with the training data and fails to generalize for the prediction of future data sets. In this study, the general trend of training accuracies is below the validation accuracies. It shows that the model does not overfit to the training data. However, this kind of trend (i.e. training accuracy < validation accuracy) is rarely observed in CNNs. A possible explanation for this behavior is the use of dropout layers. Dropout layers reduce overfitting by randomly disabling neurons, which forces the network to work on incomplete representation at subsequent layers (Srivastava et al., 2014). Dropout layers are only used during training and skipped during validation. Therefore, it is expected for dropout layers to hurt training error.

The performance of the model is further analyzed by the confusion matrix that compares the actual and the predicted labels of the validation data (Table 4).

Table 4. Confusion matrix

|  |  | Predicted | |
|---|---|---|---|
|  |  | Hazard | Non-hazard |
| **Actual** | Hazard | 53 | 13 |
|  | Non-hazard | 14 | 118 |

The confusion matrix shows that the model successfully predicts hazardous roof conditions with ~80% accuracy (53 out of 66), and successfully predicts non-hazardous roof conditions with ~89% accuracy (118 out of 132). The difference between the accuracy ratio of two classes may be due to the higher number of non-hazardous roof condition training data, resulting in the model's tendency to classify in favor of the non-hazard class.

The accuracy of the CNN classifier verifies that transfer learning is a useful approach for training a network with an input dataset that is insufficient for training from scratch. The low-level features obtained by the network trained on ImageNet data provide a suitable starting point for training the last layers of the CNN classifier which detects expert identified features of hazardous roof conditions.

To understand the logic behind the system's predictions, an integrated gradient technique is used on every image in the validation dataset. Figure 13 shows examples of the results of integrated gradients.

For hazardous roof conditions, integrated gradients show that the network's prediction is based on texture changes from light colors to darker colors. Texture changes are the results of changes in depth of roof formations where high horizontal stresses have created roof beams. For non-hazardous roof conditions, integrated gradients show that the network's prediction is based on regions where the texture colors remain stable. These regions exhibit smooth roof conditions.

Investigating the results of the integrated gradients allows the user to understand the logic behind the network's predictions. The CNN network for roof-fall-hazard detection uses pixels around the roof beams for hazardous roof condition predictions, whereas for the non-hazardous-roof-condition predictions it uses the pixels around the smooth areas. The logic used by CNN is the same as the experts' description of their intuitive decision-making skills in identifying features of hazardous and non-hazardous roof conditions. In this respect, the system mimics the expert's judgment on hazard detection. This work provides an important example of providing human-understandable justification of network predictions to create a transparent hazard management tool.

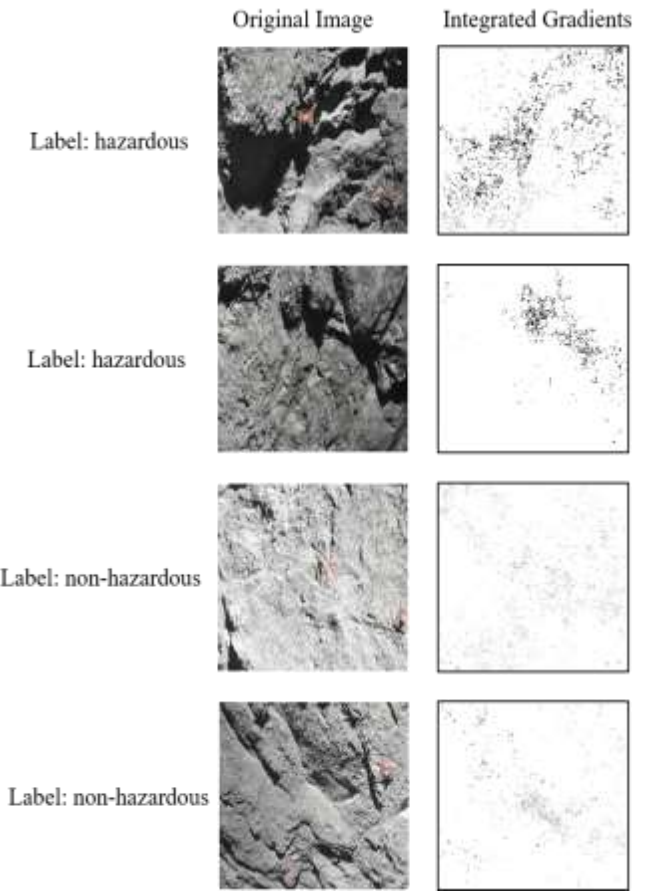

Figure 12. Results of integrated gradients method

### 5. CONCLUSIONS

Roof falls caused by high horizontal stresses are critical problems in limestone mines in the Eastern and Midwestern United States. Apart from fundamental hazard management techniques such as roof bolting and stress control layout, roof fall hazards are managed by daily visual inspections. In the Subtropolis limestone mine, these visual inspections are carried out by expert ground control personnel who look for horizontal stress-induced roof beams. In this study, a

CNN-based, autonomous roof-fall-hazard detection system was developed for the Subtropolis case. The system was trained with images showing hazardous and non-hazardous roof conditions in Subtropolis mine, followed by a transfer learning approach. Classification results are interpreted using a deep learning interpretation technique called integrated gradients. The integrated gradients show that the system mirrors expert decision-making skills on roof fall hazard detection.

The mining and tunneling industries expect the increasing use of autonomous systems. Identifying geological hazards without human involvement is an important milestone to achieve fully autonomous mining operations. This study presents the first step toward autonomous geological hazard detection in the mining and underground construction industry. Combining the network developed in this study with robotic systems or autonomous vehicles can provide continuous and real-time hazard detection.

In cases where hazard management depends on visual inspections, ground control personnel are exposed to roof fall hazards daily. Autonomous hazard detection tools eliminate these risks since human input is only required during initial system development. The role of the expert, then, becomes that of making the decisions and determining actions to be taken based on the predictions by the system. Experts are no longer exposed to roof fall safety hazards.

Intuitive skills for hazard detection are learned by experience. It means that training new personnel for roof hazard detection takes extensive effort and time. Also, knowledge may be lost when an experienced employee leaves. This brings serious safety risks to the personnel still working in the mine. The AI-based system proposed in this study achieves high-accuracy in replicating the expert judgment on hazard detection and can perform at a stable accuracy within the same geological and operational conditions. Therefore, AI-based hazard detection systems could help prevent the loss of expert knowledge that has been created through years of experience and prevents safety risk in the case of expert unavailability.

The classification capabilities of the roof fall hazard detection system developed in this study depend on ground control experts' labeling of hazardous and non-hazardous conditions. Therefore, it can be argued that the system classification accuracy can only be as good as an expert classification. However, the quantitative analysis of the relationship between the expert's labeling and roof beams showed that the expert classification is highly accurate. The system presented in this paper was developed to address the roof fall problems of a single mine. To generalize the system, the training set must be expanded with data from other sites.

Future research will focus on improving the accuracy of the algorithm by implementing networks trained with different datasets during the transfer learning stage and increasing the size of the input training dataset by rendering images from three-dimensional models of hazardous roof conditions created with digital photogrammetry.

### Acknowledgment

The research for this paper was partially supported by the National Institute for Occupational Safety and Health, contract number 0000HCCR-2019-36403. The authors would like to thank Tim Miller, East Fairfield Coal Company, for providing access to Subtropolis Mine and assistance in data collection.

### Data Availability

The data that supports the findings of this study are available from the corresponding author, upon request.

### Computer Code Availability

The Python source code of roof fall hazard detection system is available on GitHub at https://github.com/erginisleyen/Roof-fall-hazard-detection. For any question please contact at email address of corresponding author of the current manuscript.